\documentclass[conference]{IEEEtran}
\IEEEoverridecommandlockouts
\ifCLASSINFOpdf
\else
\fi

\usepackage{algorithm}
\usepackage{algpseudocode}
\usepackage{amssymb}
\usepackage{amsmath}
\usepackage{mathtools}
\usepackage{pgfplots}
\usepackage{cite}
\usepackage{graphicx}
\usepackage{subcaption}
\pgfplotsset{compat=1.18}

\begin{document}
%
\title{SODA-CitrON: Static Object Data Association by Clustering Multi-Modal Sensor Detections Online\\

}

\author{\IEEEauthorblockN{Jan Nausner}
\IEEEauthorblockA{\textit{Center for Digital Safety \& Security} \\
\textit{Austrian Institute of Technology GmbH}\\
Vienna, Austria \\
jan.nausner@ait.ac.at}
\and
\IEEEauthorblockN{Kilian Wohlleben}
\IEEEauthorblockA{\textit{Center for Digital Safety \& Security} \\
\textit{Austrian Institute of Technology GmbH}\\
Vienna, Austria \\
kilian.wohlleben@ait.ac.at}
\and
\IEEEauthorblockN{Michael Hubner}
\IEEEauthorblockA{\textit{Center for Digital Safety \& Security} \\
\textit{Austrian Institute of Technology GmbH}\\
Vienna, Austria \\
michael.hubner@ait.ac.at}
}


%


\maketitle

\begingroup
\renewcommand\thefootnote{}
\footnotetext{%
© 2026 IEEE. Accepted for the 2026 International Conference on Information Fusion (FUSION 2026).
}
\endgroup

\begin{abstract}
The online fusion and tracking of static objects from heterogeneous sensor detections is a fundamental problem in robotics, autonomous systems, and environmental mapping. Although classical data association approaches such as JPDA are well suited for dynamic targets, they are less effective for static objects observed intermittently and with heterogeneous uncertainties, where motion models provide minimal discriminative power with respect to clutter. In this paper, we propose a novel method for static object data association by clustering multi-modal sensor detections online (SODA-CitrON), while simultaneously estimating positions and maintaining persistent tracks for an unknown number of objects. The proposed unsupervised machine learning approach operates in a fully online manner and handles temporally uncorrelated and multi-sensor measurements. Additionally, it has a worst-case loglinear complexity in the number of sensor detections while providing full output explainability. We evaluate the proposed approach in different Monte Carlo simulation scenarios and compare it against state-of-the-art methods, including POM-based filtering, DBSTREAM clustering, and JPDA. The results demonstrate that SODA-CitrON consistently outperforms the compared methods in terms of F1 score, position RMSE, MOTP, and MOTA in the static object mapping scenarios studied.
\end{abstract}


%
\IEEEpeerreviewmaketitle

\section{Introduction}\label{sec:introduction}

The fusion and association of object-level information from heterogeneous sensors is a central problem in robotics, autonomous systems, and surveillance. Today, perception systems increasingly rely on multi-modal sensing techniques to reliably operate in complex and uncertain environments. In a multitude of application domains, the objective is not solely to detect objects but to construct and sustain a persistent spatial map of static objects.

Such requirements emerge in a broad spectrum of use cases, including robotic exploration \cite{hroob_adaptive_2024}, environmental and agricultural mapping \cite{samadzadegan_critical_2025}, and autonomous driving \cite{shi_radar_2026}. Here, it is vital to detect and track static landmarks such as trees, crops, terrain features, roadside structures, and other persistent environmental objects to perform navigation and maintain long-term autonomy. Similar demands arise in search \& rescue, survey, and threat detection scenarios. In these situations, static objects that pose a potential threat to humans (e.g., chemical, biological, radiological, nuclear, and explosive (CBRNE) hazards) must be reliably detected and precisely located \cite{schraml2022real, hasselmann_multi-robot_2024, lekhak_viability_2024}. Furthermore, the ability to consistently associate repeated detections from heterogeneous sensors with persistent object hypotheses is critical, especially in the context of emergency response. This is crucial for minimizing the risk of loss of life and human exposure to danger and for enabling informed decision-making during operations.

In this paper, we explicitly focus on online data association for static objects, observed asynchronously by heterogeneous sensors. Our proposed approach, termed SODA-CitrON, employs clustering of multi-modal detections for concurrently estimating positions and maintaining persistent tracks for an unknown number of objects, while reducing clutter and providing output explainability. The novel method runs fully online with worst-case loglinear complexity in the number of detections, rendering it suitable for real-time systems processing large numbers of detections.

\subsection{Related Work}

Data association has historically been explored in the context of multi-target tracking, where classical approaches, including joint probabilistic data association (JPDA) and its variants, are well established and highly effective for tracking moving targets observed at regular intervals \cite{blackman1999design}. However, they are considerably less suitable for situations with mainly static objects, where motion models provide little discriminative power, observations may be temporally sparse or uncorrelated, and sensor-specific uncertainties dominate the association problem \cite{bar1990tracking, blackman1999design}. In such cases, the association process must rely primarily on spatial proximity, measurement uncertainty, and sensor characteristics. Several works explicitly addressing static object association \cite{barker_static_1998, guler_stationary_2007, schueler_360_2012, bowman_probabilistic_2017} remain limited in terms of multi-modal detection handling, positioning accuracy, and online capability.

This becomes even more challenging in the context of heterogeneous sensor setups, where differing resolutions, fields of view, noise characteristics, and detection semantics must be considered. Bayesian filters operating on probabilistic occupancy maps (POMs) \cite{hubner2024bayesian, wohlleben_bayesian_2025} aim to improve feature-level fusion in mixed sensor systems. They focus on reducing false positives and increasing robustness, but lack detection association mechanisms. Additionally, clustering-based approaches have been investigated for data association, track-to-track fusion, and multi-target tracking \cite{s21175715, 8999712, nurfalah_effective_2024}. These approaches typically use density-based algorithms
and focus on moving objects. Online clustering methods for data streams \cite{cao_density-based_2006, hahsler_clustering_2016} provide a solid foundation for associating repeated detections due to their scalability and ability to handle an unknown number of objects. However, they lack mechanisms for maintaining persistent object identities and explicit modeling of heterogeneous sensor uncertainties and detection confidences.

\subsection{Contributions}
In this work, a novel method, named SODA-CitrON, is proposed to perform static object data association in clutter. SODA-CitrON is rooted in the density-based online clustering method DBSTREAM \cite{hahsler_clustering_2016}, which is extended with the following mechanisms:

\begin{itemize}
    \item Non-linear confidence-based weighting of sensor detections for static object track initiation
    \item Information filtering for online static object state estimation
    \item Unique ID assignment for consistent object tracking
\end{itemize}

Furthermore, the computational complexity of SODA-CitrON is analyzed. Finally, a Monte Carlo simulation framework for systematically evaluating the proposed methodology and related state-of-the-art methods on heterogeneous multi-sensor, multi-object static object data association scenarios is introduced.

\subsection{Paper Organization}
The structure of this paper is as follows. Section \ref{sec:methodology} provides a detailed description of the proposed methodology, starting with the problem statement and finalizing with a summary of key assumptions and a computational complexity analysis. Section \ref{sec:evaluation} elaborates on the evaluation setup, encompassing the experimental configuration, the scenarios considered, and the methods that were utilized for comparison. In Section \ref{sec:results}, the results are discussed and limitations analyzed. Finally, section \ref{sec:conclusion} concludes the paper by summarizing the key takeaways and outlining directions for future work.

\section{Methodology}\label{sec:methodology}

\subsection{Problem Statement}\label{sec:problem-statement}

Multiple static (non-moving) objects $k \in \mathcal{K} = \{1,\dots,N_K\}$, potentially of different type ($t_k\in\mathcal{T}$), with unknown true positions $\mathbf{x}_k = [x_k, y_k]^T$ are spread over a region of interest (ROI). Multiple sensors $s \in \mathcal{S} = \{1,\dots,N_S\}$, stationary or moving, scan the ROI in order to detect all objects in $\mathcal{K}$. Each sensor is capable of detecting a subset of all object types, with varying detection probabilities for each type ($P_{D}^{(t,s)}, t \in \mathcal{T}, s \in \mathcal{S}$). All sensors together produce a sequence of $N_Z$ detections $Z_n^{(s)} = (\mathbf{z}_n^{(s)},\ \pi_n^{(s)},\ \mathbf{R}_n^{(s)}),\ s \in \mathcal{S},\ n \in \{1,\dots,N_Z\}$, with measured positions $\mathbf{z}_n^{(s)} = [x_n^{(s)}, y_n^{(s)}]^T$, confidence (probability of true object detection) $\pi_n^{(s)} \in [0,1]$, which is typically related to the signal-to-noise ratio (SNR) at the detector, and position covariance matrix $\mathbf{R}_n^{(s)}\in\mathbb{R}^{2\times2}$. Each of these detections is a true object detection (with probability $\pi_n^{(s)}$) or clutter (with probability $1-\pi_n^{(s)}$). Sensors are not assumed to perform simultaneous ROI scans. Hence, the detection sequence can be of arbitrary order and the temporal correlation of detections of a particular object by different sensors is not presupposed.

For each sensor $s\in\mathcal{S}$, the detected position of the object $\mathbf{x}_k,\ k\in\mathcal{K}$ is given by
\begin{equation}\label{eq:measurement-model}
    \mathbf{z}_n^{(s)} = \mathbf{H}^{(s)}\mathbf{x}_k+\mathbf{v}_n^{(s)}\text{, }\mathbf{v}_n^{(s)}\sim\mathcal{N}(\mathbf{0},\mathbf{R}_n^{(s)})\text{,}
\end{equation}
where $\mathbf{H}^{(s)}\in\mathbb{R}^{2x2}$ represents the known measurement model and $\mathbf{v}_k^{(s)}$ specifies the measurement noise, defined by a zero-mean gaussian distribution with known (estimated) noise covariance $\mathbf{R}_n^{(s)}$.

The goal is to estimate the set of objects $\hat{\mathcal{K}}=\mathcal{C}_n$, based on the sequence of sensor detections $Z_n^{(s)},\ s \in \mathcal{S},\ n \in \{1,\dots,N_Z\}$. The estimated number of objects present in the ROI is thus given by $\hat{N_K}=|\mathcal{C}_n|$. Each object $c\in\mathcal{C}_n$ shall be defined by its estimated position $\hat{\mathbf{x}}_{n}^{(c)}$. Furthermore, a position covariance $\mathbf{P}_{n}^{(c)}$ should be provided. 
As indicated by the index $n$, this computation should be performed online: the method must be able to provide intermediate state estimates based on the detections it has seen so far, not only in the end, when all detections are available.

\subsection{Sequential State Estimation for Static Objects}

First, the problem of sequentially updating the estimated position $\hat{\mathbf{x}}_{n}^{(c)},\ n \in \{1,\dots,N_Z\}$ of an object $c\in\mathcal{C}$ given sensor detection tuples $Z_{i}^{(s)},\ i \in \{1,\dots,n\}$ is addressed in isolation. The joint static object data association and state estimation will be further developed in Section \ref{sec:static-object-data-association}. Based on the assumptions stated in Section \ref{sec:problem-statement}, the information filter is a good choice to solve this problem \cite{bar2001estimation}. The information filter is an algebraically equivalent formulation of the Kalman filter, centered around the information matrix 
\begin{equation}\label{eq:information-matrix}
    \mathbf{Y}_{n}^{(c)} = {\mathbf{P}_{n}^{(c)}}^{-1}\in\mathbb{R}^{2\times2}
\end{equation}
as the inverse of the posterior covariance matrix $\mathbf{P}_{n}^{(c)}$ and the information vector
\begin{equation}\label{eq:infofilter-state}
    \hat{\mathbf{y}}_{n}^{(c)} = {\mathbf{P}_{n}^{(c)}}^{-1}\hat{\mathbf{x}}_{n}^{(c)}
\end{equation}
respectively. Using this formulation allows for a simplification of the filter update step, as compared to the Kalman filter. Note that the filter prediction step is not relevant for static targets and hence can be omitted. The update step becomes a simple sum for the information matrix
\begin{equation}\label{eq:infofilter-update-Y}
\begin{split}
    \mathbf{Y}_{n}^{(c)} &= \mathbf{Y}_{n-1}^{(c)} + \mathbf{H}^{(s)}{\mathbf{R}_n^{(s)}}^{-1}{\mathbf{H}^{(s)}}^T\\
    &= \mathbf{Y}_{0}^{(c)} + \sum_{i=1}^{n}\mathbf{H}^{(s)}{\mathbf{R}_i^{(s)}}^{-1}{\mathbf{H}^{(s)}}^T
\end{split}
\end{equation}
as well as for the information vector
\begin{equation}\label{eq:infofilter-update-y}
\begin{split}
    \hat{\mathbf{y}}_{n}^{(c)} &= \hat{\mathbf{y}}_{n-1}^{(c)} + \mathbf{H}^{(s)}{\mathbf{R}_n^{(s)}}^{-1}\mathbf{z}_n^{(s)}\\
    &= \hat{\mathbf{y}}_{0}^{(c)} + \sum_{i=1}^{n}\mathbf{H}^{(s)}{\mathbf{R}_i^{(s)}}^{-1}\mathbf{z}_i^{(s)}\text{.}
\end{split}
\end{equation}
The information filter is initialized with $\mathbf{Y}_{0}^{(c)}=\mathbf{0}$ and $\hat{\mathbf{y}}_{0}^{(c)}=\mathbf{0}$. The posterior covariance and state estimates can be obtained at any time step by $\mathbf{P}_{n}^{(p)} = {\mathbf{Y}_{n}^{(p)}}^{-1}$ and $\hat{\mathbf{x}}_{n}^{(p)} = \mathbf{P}_{n}^{(p)}\hat{\mathbf{y}}_{n}^{(p)}$ respectively.

Furthermore, the information filter formulation allows for simple track-to-track fusion of objects. The information matrix of fused object $c'$, resulting from the fusion of objects $c_1, c_2\in\mathcal{C},\ c_1 \neq c_2$, is given by
\begin{equation}\label{eq:infofilter-merge-covariance}
    \mathbf{Y}_{n}^{(c')} = \mathbf{Y}_{n}^{(c_1)}+\mathbf{Y}_{n}^{(c_2)}
\end{equation}
and the fused information vector is obtained as
\begin{equation}\label{eq:infofilter-merge-state}
    \hat{\mathbf{y}}_{n}^{(c')} = \hat{\mathbf{y}}_{n}^{(c_1)}+\hat{\mathbf{y}}_{n}^{(c_2)}\text{.}
\end{equation}

\subsection{Static Object Data Association}\label{sec:static-object-data-association}

In the next step, the problem of online static object data association and state estimation with appropriate track initiation is addressed. This essentially amounts to deciding for each new sensor measurement if it is likely a true object detection and later, to which object it should be associated, or if it is clutter (known as the data association problem). Furthermore, a mechanism is required to initialize tracked objects from the sensor measurements seen so far (known as the track initiation problem). For this purpose, the state-of-the-art density-based online clustering algorithm DBSTREAM \cite{hahsler_clustering_2016} is extended. This method is well suited to solve the problem introduced in Section \ref{sec:problem-statement}, as detections are expected to form dense clusters around the true object positions, while clutter is expected to spread uniformly with low density throughout the ROI. Furthermore, it has the advantage of being able to process data streams sequentially, as opposed to related traditional clustering methods such as DBSCAN \cite{ester_density-based_1996}. The resulting novel method is named SODA-CitrON. Note that in the following paragraphs the indices $n$ and $s$ will be omitted to increase readability, and $\mathbf{H}^{(s)} = \mathbf{I},\ \forall s \in \mathcal{S}$.

At the core of SODA-CitrON is the set of potential objects $\mathcal{P} = \{1,\dots,N_P\}$ (referred to as micro-clusters in \cite{hahsler_clustering_2016}), where $\mathcal{C} \subseteq \mathcal{P}$ (the set of estimated objects $\mathcal{C}$ is referred to as clusters in \cite{hahsler_clustering_2016}). The initialization of a new potential object is shown in Lines 5-7 of Algorithm \ref{alg:soda-citron-update}, starting with the creation of a new unique object ID. 
The state of any potential object $p \in \mathcal{P}$ is given by $P^{(p)} = (\hat{\mathbf{y}}^{(p)}, \mathbf{Y}^{(p)}, w^{(p)})$, including information filter states and confidence-based weight, which will be introduced in the following paragraphs.

\begin{figure}
    \centering
    \begin{tikzpicture}
    \begin{axis}[
        width=\linewidth*0.9,
        height=4cm,
        domain=0:1,
        samples=200,
        xlabel={$x$},
        ylabel={$f(x)$},
        xmin=0, xmax=1,
        ymin=0, ymax=10,
        grid=both,
        axis lines=left,
        legend style={
            at={(0.05,0.95)},
            anchor=north west,
            draw=none
        }
    ]

    \addplot[
        thick,
        dashed
    ]
    {10*(exp(3*x)-1)/(exp(3)-1)};
    \addlegendentry{$\beta=3$}
    \addplot[
        thick,
    ]
    {10*(exp(6*x)-1)/(exp(6)-1)};
    \addlegendentry{$\beta=6$}
    \addplot[
        thick,
        dotted
    ]
    {10*(exp(9*x)-1)/(exp(9)-1)};
    \addlegendentry{$\beta=9$}
    
    \end{axis}
    \end{tikzpicture}
    \caption{Confidence to weight transformation (Eq. \ref{eq:confidence-weight-transformation}) for $\beta \in \{3,6,9\}$ and $w_{max}=10$.}
    \label{fig:confidence-to-weight}
\end{figure}
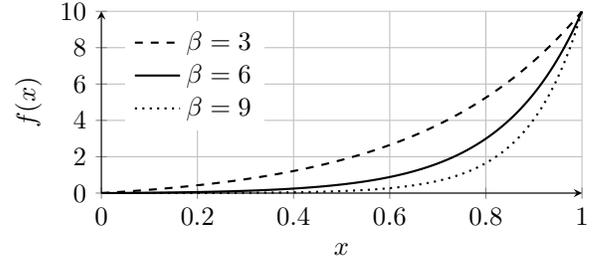

In a first extension of DBSTREAM, SODA-CitrON introduces the confidence-based weight $w$ for a detection $Z$, as shown in Line 3 of Algorithm \ref{alg:soda-citron-update}. State-of-the-art tracking systems often use track initiation rules based on the number of associated detections (e.g. detections in N out of M time steps) or based on the track score \cite{blackman1999design}, which can be well suited for moving objects, but were found to be over-confident for the static object situation presented in Section \ref{sec:problem-statement}. To address this issue, the detection weight is computed via a non-linear transformation based on the detection confidence $\pi$:
\begin{equation}\label{eq:confidence-weight-transformation}
    f(x) \coloneq \frac{e^{\beta x}-1}{e^{\beta}-1}w_{max},\ \forall x\in[0,1]\text{, with }w = f(\pi)\text{.}
\end{equation}
The transformation $f(x)$ is parameterized with the steepness factor $\beta$ and maximum weight $w_{max}$, and different parameterizations of $f(x)$ are shown in Fig. \ref{fig:confidence-to-weight}. This weight transformation strongly down-weighs low confidence detections, leading to a situation where a disproportionately high number of low confidence detections is required to initiate a target, while conversely only a single high-confidence detection suffices to initiate an object. If a detection is associated with a potential object, the detection weight is added to the weight $w^{(p)}$ of the potential target, as shown in Line 12 of Algorithm \ref{alg:soda-citron-update}. Note that the actual object (track) initiation occurs in Line 19 of Algorithm \ref{alg:soda-citron-reclustering}, explained in more detail in the following paragraphs.

Furthermore, SODA-CitrON introduces an information filter in DBSTREAM to sequentially update the position estimate and position covariance of a potential object $p \in \mathcal{P}$. If a new sensor detection $Z$ is associated with an existing potential object, the filter is updated accordingly with the measured position and covariance, as shown in lines 10-11 in Algorithm \ref{alg:soda-citron-update}, implementing Eqs. \ref{eq:infofilter-update-Y} and \ref{eq:infofilter-update-y}. 
Note that the remaining lines 13-25 of Algorithm \ref{alg:soda-citron-update} remain unchanged compared to DBSTREAM, leaving the shared density update and cluster collapse prevention mechanisms intact.

\begin{algorithm}[t]
\caption{SODA-CitrON Update}
\label{alg:soda-citron-update}
\begin{algorithmic}[1]
    \Require Input: $Z=(\mathbf{z}, \pi, \mathbf{R})$
    \Require Parameters: $r$, Transformation: $f(x)$
    \Require Internal State: $\mathcal{P}$, $\mathcal{P}_{0} \gets \emptyset$, $\mathcal{D}$, $\mathcal{D}_{0} \gets \emptyset$
    \State $\mathcal{N} \gets \{p \in \mathcal{P}\ |\ ||\hat{\mathbf{x}}^{(p)}-\mathbf{z}|| < r\}$ \Comment{Determine neighbors}
    \State $\mathbf{Y} \gets \mathbf{R}^{-1}$ \Comment{Input information matrix}
    \State $w \gets f(\pi)$ \Comment{Confidence-based weight}
    \If{$|\mathcal{N}| < 1$} \Comment Add new potential object
        \State $p \gets$ newID()
        \State $\mathcal{P} \gets \mathcal{P} \cup \{p\}$, $\hat{\mathbf{y}}^{(p)} \gets \mathbf{Y}\mathbf{z}$, $\mathbf{Y}^{(p)} \gets \mathbf{Y}$
        \State $l^{(p)} \gets \log\frac{\pi}{1-\pi}$, $w^{(p)} \gets w$
    \Else \Comment Update existing potential objects
        \For{$i \in \mathcal{N}$}
            \State $\hat{\mathbf{y}}^{(i)} \gets \hat{\mathbf{y}}^{(i)} + \mathbf{Y}\mathbf{z}$
            \State $\mathbf{Y}^{(i)} \gets \mathbf{Y}^{(i)} + \mathbf{Y}$
            \State $w^{(i)} \gets w^{(i)} + w$
            \For{$j \in \mathcal{N}$ and $j > i$} \Comment{Update shared density}
                \State $d^{(ij)} \gets d^{(ij)} + w$
                \If{$d^{(ij)} \not\in \mathcal{D}$}
                    \State $\mathcal{D} \gets \mathcal{D} \cup \{d^{(ij)}\}$
                \EndIf
            \EndFor
        \EndFor
        \For{$(i,j) \in \mathcal{N}\times\mathcal{N}$ and $i > j$}
            \If{$||\hat{\mathbf{x}}^{(i)}-\hat{\mathbf{x}}^{(j)}|| < r$} \Comment{Prevent collapse}
                \State Revert $P^{(i)}, P^{(j)}$ to previous states
            \EndIf
        \EndFor
    \EndIf
\end{algorithmic}
\end{algorithm}

The reclustering step of SODA-CitrON is essentially the same as in DBSTREAM, preserving the search for connected components within the connectivity graph, given by the set of potential targets $\mathcal{P}$ and the weighted adjacency list $\mathcal{V}$. The intersection factor is set to $\alpha=0.3$, as suggested in \cite{hahsler_clustering_2016}. The main addition is in the adaption of the potential object fusion. Lines 12-14 in Algorithm \ref{alg:soda-citron-reclustering} show the fusion of extended potential object states according to the methods outlined in Section \ref{sec:problem-statement}. The final objects $\mathcal{C}$ are determined after reclustering by selecting all potential objects that exceed the minimum weight $w_{min}$. This is shown on Line 19 of Algorithm \ref{alg:soda-citron-reclustering}. Recall that the object position estimates and position covariances can be recovered from $P^{(c)},\ c\in\mathcal{C}$ according to the relations shown in Eqs. \ref{eq:information-matrix} and \ref{eq:infofilter-state}.
The parameter $w_{min}$ mainly affects the clutter filtering ability, where a large $w_{min}$ tends to decrease the number of false positives at the cost of fewer true positives, and a small $w_{min}$ tends to show the opposite effect. The clustering threshold $r$ favors the separation of closely spaced objects when set to a small value, at the cost of potentially rejecting valid neighboring detections. A large $r$ in turn promotes the merging of closely spaced objects.

Finally, note that the fading and cleanup mechanism of DBSTREAM is removed from SODA-CitrON because by definition there is no movement among static objects.

\begin{algorithm}[t]
\caption{SODA-CitrON Reclustering}
\label{alg:soda-citron-reclustering}
\begin{algorithmic}[1]
    \Require Parameters: $w_{min}, \alpha$
    \Require Internal State: $\mathcal{P}$, $\mathcal{D}$
    \Ensure Output: $\mathcal{C}$
    \State $\mathcal{V} \gets \emptyset$ \Comment{Weighted adjacency list}
    \For{$d^{(ij)} \in \mathcal{D}$}
        \If{$w^{(i)} \geq w_{min}$ and $w^{(j)} \geq w_{min}$}
            \State $v^{(ij)} \gets \frac{d^{(ij)}}{(w^{(i)}+w^{(j)})/2}$
            \State $\mathcal{V} \gets \mathcal{V} \cup \{v^{(ij)}\}$
        \EndIf
    \EndFor
    \State $\mathcal{G} \gets$ connectedComponents($\mathcal{V} \geq \alpha$)
    \For{$g \in \mathcal{G}$}
        \For{$i \in g$}
            \For{$j \in g$ and $i > j$} \Comment{Fuse connected objects}
                \State $\hat{\mathbf{y}}^{(i)} \gets \hat{\mathbf{y}}^{(i)} + \hat{\mathbf{y}}^{(j)}$
                \State $\mathbf{Y}^{(i)} \gets \mathbf{Y}^{(i)} + \mathbf{Y}^{(j)}$
                \State $w^{(i)} \gets w^{(i)} + w^{(j)}$
                \State $\mathcal{P} \gets \mathcal{P} \setminus \{j\}$ \Comment{Delete fused object}
            \EndFor
        \EndFor
    \EndFor
    \State $\mathcal{C} \gets \{p \in \mathcal{P}\ |\ w^{(p)} \geq w_{min}\}$ \Comment{Determine final objects}
\end{algorithmic}
\end{algorithm}

\subsection{Key Assumptions}

SODA-CitrON relies on the following assumptions in order to perform well in static object data association scenarios with heterogeneous multi-sensor systems:

\begin{itemize}
    \item Sensor detection position errors are moderate to low.
    \item Sensors provide at least one high-confidence detection or multiple low confidence detections of an object.
    \item Sensors provide a meaningful confidence estimate for their detections.
    \item Clutter detections generally have lower detection confidence than true detections and are more sparse.
\end{itemize}

\subsection{Computational Complexity}

In \cite{hahsler_clustering_2016} it is shown that the time complexity for the clustering (update) step is in $\mathcal{O}(dN_Z\log (|\mathcal{P}_{max}|)+N_Z|\mathcal{N}_{max}|^2)$ and $\mathcal{O}(|\mathcal{P}_{max}| \log (|\mathcal{P}_{max})|) + 2|\mathcal{P}_{max}||\mathcal{N}_{max}| + |\mathcal{P}_{max}|)$ for the reclustering step. $d$ refers to the dimensionality of the vectors $\mathbf{x}$ and $\mathbf{z}$, where setting the dimensionality $d=2$ yields $\mathcal{N}_{max}=6$ \cite{hahsler_clustering_2016}. Furthermore, the worst case number of potential objects $|\mathcal{P}_{max}|$ is equal to the total number of detections $N_Z$, representing a situation where there are no connected objects available for fusion during the reclustering step. Using this knowledge, the complexities are updated to $\mathcal{O}(2N_Z\log(N_Z)+36N_Z)$ for the clustering step and to $\mathcal{O}(N_Z\log (N_Z) + 12N_Z + N_Z)$ for the reclustering step, yielding a final simplified asymptotic worst case time complexity of $\mathcal{O}(N_Z\log(N_Z))$ for SODA-CitrON. Note that this result is different from \cite{hahsler_clustering_2016}, as the fading and cleanup mechanisms are not used in SODA-CitrON. This time complexity analysis only holds under the assumption that the neighborhood computation in line 1 of listing \ref{alg:soda-citron-update} is implemented efficiently, e.g., by using spatial indexing data structures such as R-trees \cite{guttman_r-trees_1984}. In the common case with medium clutter and many connected components in the shared density graph $|\mathcal{P}_{max}|$ is expected to be much smaller than $N_Z$, implying even faster computation.

\section{Evaluation}\label{sec:evaluation}

\subsection{Setup}

In this work, evaluation of SODA-CitrON and comparison to other methods are done via Monte Carlo simulations. For this purpose, a set of object types $\mathcal{T} = \{A, B, C, D\}$ is defined. 
Furthermore, the set of simulated sensors $\mathcal{S} = \{S1, S2, S3, S4, S5\}$ is defined, with the relevant sensor parameters listed in Tab. \ref{tab:sensor_params}. Note that $\mathcal{N}$ signifies a discrete normal distribution, $B$ signifies the beta distribution, and  $\pi_{S1}(x)$ represents the following probability mass function:
\begin{equation}
    \pi_{S1}(x) = \begin{cases}
    0.25 & x = 0.5 \vee x = 0.75 \\
    0.5 & x = 1.0 \\
    0 & \text{otherwise}
    \end{cases} \text{.}
\end{equation}
In Tab. \ref{tab:sensor_params}, $P_D^{(s,t)}, s \in \mathcal{S}, t \in \mathcal{T}$ is the detection probability of sensor type $s$ with respect to object type $t$. The parameter Detections/object determines the number of detections generated by a sensor given a specific object type. The sensor position covariance matrix is given by $\mathbf{R}^{(s)} = diag((\sigma^{(s)})^2,(\sigma^{(s)})^2), s \in \mathcal{S}$. Detection and clutter confidence parameterize the simulated sensor confidence of true and clutter detections, respectively. Finally, the clutter rate (clutter detections / m$^2$) determines the poisson-distributed number of clutter detections per sensor, which are placed uniformly over the ROI.
\begin{table}[h]
\caption{Sensor ($S1$--$S5$) simulation parameters.}
\label{tab:sensor_params}
\centering
\resizebox{\columnwidth}{!}{%
\begin{tabular}{l|ccccc}
\hline
\textbf{Parameter} & $S1$ & $S2$ & $S3$ & $S4$ & $S5$ \\
\hline
$P_D^{(s,A)}$ & 0.4 & 0.8 & - & 0.6 & 0.8 \\
$P_D^{(s,B)}$ & 0.7 & - & 0.85 & 0.6 & 0.3 \\
$P_D^{(s,C)}$ & 0.9 & 0.4 & 0.4 & 0.6 & 0.7 \\
$P_D^{(s,D)}$ & 0.8 & 0.4 & 0.4 & 0.6 & 0.7 \\
Detections/object ($A$) & 1 &$\mathcal{N}(3,1)$& - & 1 & $\mathcal{N}(2,1)$ \\
Detections/object ($B$) & 1 & - & 1 & 1 & $\mathcal{N}(2,1)$ \\
Detections/object ($C$) & 1 &$\mathcal{N}(3,1)$& 1 & 1 & $\mathcal{N}(2,1)$ \\
Detections/object ($D$) & 1 &$\mathcal{N}(3,1)$& 1 & 1 & $\mathcal{N}(2,1)$ \\
\hline
Position covariance $(\sigma^{(s)})^2$ & 0.015 & 0.167 & 0.082 & 0.082 & 0.376 \\
Detection confidence $\pi_{det}^{(s)}$ & $\pi_{S1}(x)$ & $B(8, 2.5)$ & $B(8, 2.5)$ & $B(8, 2.5)$ & $B(8, 2.5)$ \\
Clutter confidence $\pi_{clutter}^{(s)}$ & $\pi_{S1}(x)$ & $B(8, 8)$ & $B(8, 8)$ & $B(8, 8)$ & $B(8, 8)$ \\
Clutter rate $\lambda^{(s)}$ & 0.0005 & 0.02 & 0.01 & 0.01 & 0.02 \\ 
\hline
\end{tabular}
}
\vspace{-1.5mm}
\end{table}
All Monte Carlo experiments were repeated 500 times. For the simulations, a computation node with 2 AMD EPYC 9254 24-Core processors and 512 GiB RAM was used.

\subsection{Scenarios}

\begin{figure*}[h]
  \centering
  \includegraphics[width=\textwidth, trim = 0.18cm 0.3cm 0.05cm 0.15cm, clip]{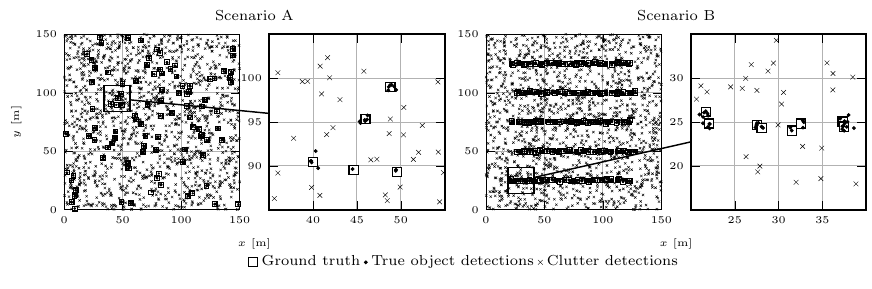}
  \caption{One Monte Carlo instance of scenario A and scenario B, with ground truth and simulated sensor detections.}
  \label{fig:scenarios}
\end{figure*}
Two simulation scenarios (A and B) were defined for the Monte Carlo experiments, as visualized in Fig. \ref{fig:scenarios}. These scenarios and the respective sensor simulation parameters in Tab. \ref{tab:sensor_params} were crafted according to the requirements from the search \& rescue and CBRNE threat mapping use cases, as introduced in Section \ref{sec:introduction}. In scenario A, 25 objects of each object type (A,B,C,D) are randomly placed on an ROI of 150 m x 150 m, following a uniform distribution. For scenario B, objects of type A are placed in 5 rows (inter-row distance ca. 25 m) with spacing of ca. 5 m within rows (object positions are sampled randomly with mild noise) over the same ROI. Next to each object of type A, an object of type B is randomly placed within a radius of 0.5-1.5 m. In total, there are 105 objects of type A and 105 objects of type B in the scenario. Scenario A had between 1657 and 1897 sensor detections for each run and Scenario B had between 2060 and 2300.

\subsection{Method Comparison}

In order to compare the static object data association performance of SODA-CitrON, the following methods with their respective parameters were selected.
\subsubsection{POM-based filter \cite{hubner2024bayesian}} Bayesian POM fusion with threshold $\theta = 0.75$, resolution 0.1 m x 0.1 m, no decay and no prior information.
\subsubsection{JPDA \cite{blackman1999design}} Implementation of \cite{10224185} with measurement noise = 0.12, clutter rate = 0.01, $P_D=0.9$, $P_G=0.9$, missed distance = 5.0 and track initiator with minimum detections = 3. The chosen motion model is an almost static random walk model with a noise diffusion coefficient = 0.0001 for both dimensions.
\subsubsection{DBSTREAM \cite{hahsler_clustering_2016}} Implementation of \cite{montiel_river_2020}
with $r=1.1$, $\lambda=0.0$, $t_{gap}=1$, $w_{min}=3$ and $\alpha=0.3$.
\subsubsection{SODA-CitrON} With $\beta=6$, $w_{max}=10$, $r=1.1$, $w_{min}=4.0$ and $\alpha=0.3$.

Parameter selection is based on the scenario requirements and several test trials, which yielded a favorable parameter set for each method. Note that systematic optimization of algorithmic parameters is outside the scope of this work.

\subsection{Evaluation Method}
\begin{table}[t]
\caption{Detection radius per object type ($A$--$D$).}
\label{tab:object_params}
\centering
\footnotesize
\begin{tabular}{l|cccc}
\hline
\textbf{Parameter} & $A$ & $B$ & $C$ & $D$ \\
\hline
Detection radius (normal) [m] & 0.8 & 0.7 & 0.75 & 0.95 \\
Detection radius (strict) [m] & 0.3 & 0.2 & 0.25 & 0.45 \\
\hline
\end{tabular}%
\end{table}
The different algorithms are evaluated primarily with respect to the F1 Score and the root mean squared error (RMSE) of the estimated positions. Each type of object is characterized by a detection radius as shown in Tab. \ref{tab:object_params}. This radius determines whether an estimated object, generated as an output of any static object data association method, is to be valued as a true positive or a false positive. The RMSE is then computed between the ground truths and the corresponding true detections, which are determined according to either the normal or the strict detection radius. The performance of the different algorithms over time is analyzed based on the F1 score, the multiple object tracking precision (MOTP), and the multiple object tracking accuracy (MOTA) metrics \cite{bernardin2008evaluating}. Note that only JPDA and SODA-CitrON support tracking in the sense that they assign an ID to each object, hence MOTA, incorporating ID switches, can only be computed for them.

\section{Results}\label{sec:results}

\begin{figure*}[t]
  \centering
  \includegraphics[width=\textwidth, trim = 0.4cm 0.45cm 0.4cm 0.37cm, clip]{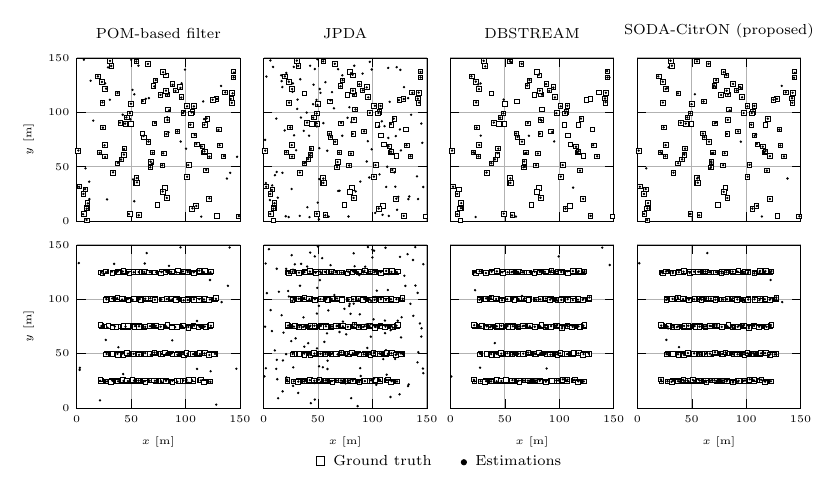}
  \caption{Resulting object position estimations from the data shown in Fig. \ref{fig:scenarios}. Top row: scenario A, bottom row: scenario B.}
  \label{fig:scenario_estimations}
\end{figure*}

Example output position estimations for each algorithm for both scenarios are shown in Fig. \ref{fig:scenario_estimations}. It is immediately apparent that, for the POM-based filter and for JPDA, more false positives are present after the data fusion step, as compared to DBSTREAM and SODA-CitrON. In both scenarios, a high detection rate is visible, likely because the runs contained a lot of sensor detections and there were detections present for most targets. Some targets could not be detected by any of the algorithms, for example the one at (73, 15), indicating that in this case there might not have been enough correlated sensor information present to reliably discern them from clutter.
\begin{table}[h]
    \caption{Average runtime [s] of the different methods.}
    \label{tab:runtimes}
    \centering
    \footnotesize
    \begin{tabular}{l|cc}
        \hline
        \textbf{Method} & Scenario A & Scenario B \\
        \hline
        POM-based filter & 746  & 1115 \\
        JPDA            & 1400 & 1978 \\
        DBSTREAM        & 36.4   & 50.0   \\
        \textbf{SODA-CitrON (proposed)}     & \textbf{6.4} & \textbf{8.4} \\
        \hline
    \end{tabular}
\end{table}
In Tab. \ref{tab:runtimes} the average runtime for each of the algorithms for both scenarios is shown. The POM-based filter and JPDA have both had a comparatively long runtime across the board, whereas DBSTREAM and SODA-CitrON have both been relatively fast.
\begin{figure*}[htbp]
     \centering
     \begin{subfigure}[b]{0.48\textwidth}
         \centering
         \includegraphics[width=\linewidth, height=4cm]{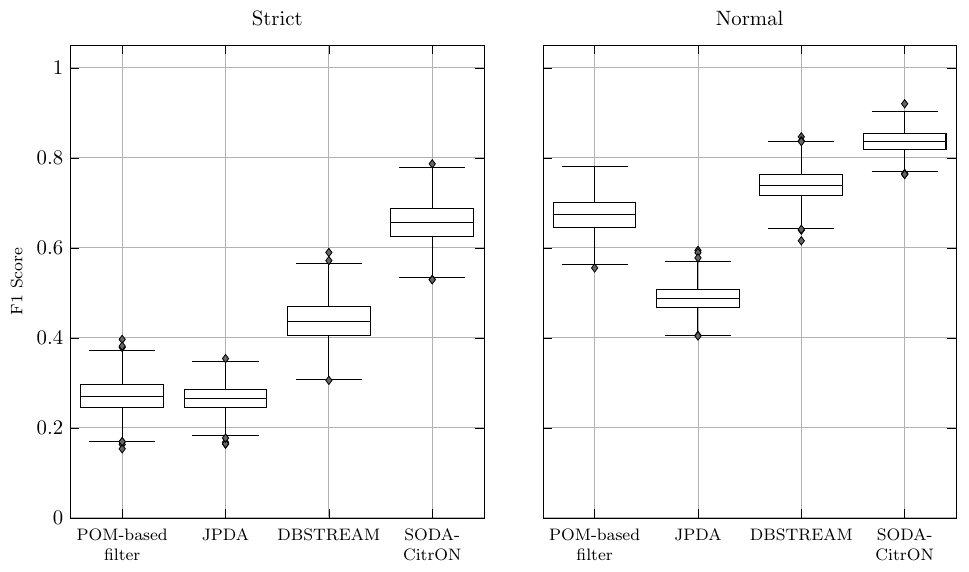}
         \caption{F1 score scenario A}
         \label{fig:F1 A}
     \end{subfigure}
     \hfill
     \begin{subfigure}[b]{0.48\textwidth}
         \centering
         \includegraphics[width=\linewidth, height=4cm]{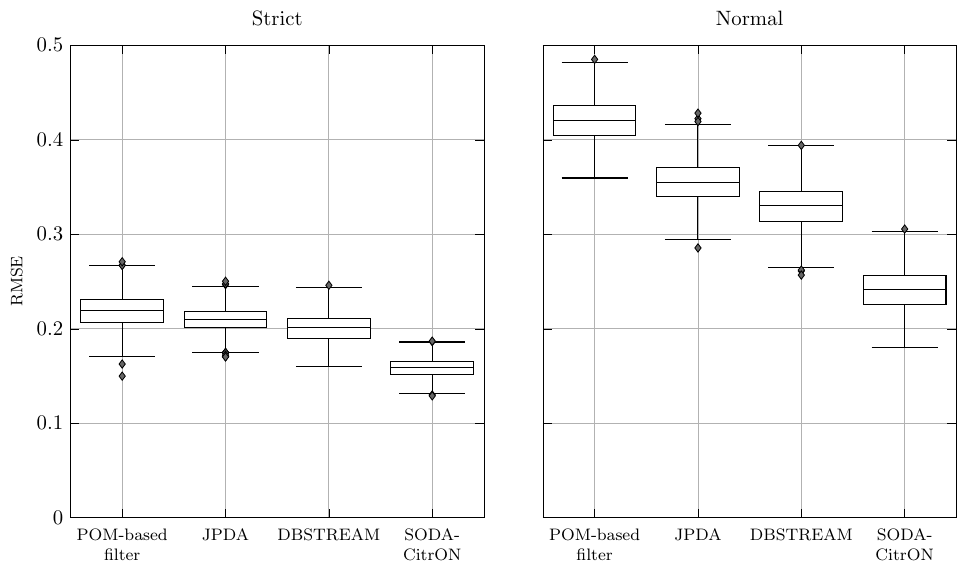}
         \caption{RMSE scenario A}
         \label{fig:RMSE A}
     \end{subfigure}

     \begin{subfigure}[b]{0.48\textwidth}
         \centering
         \includegraphics[width=\linewidth, height=4cm]{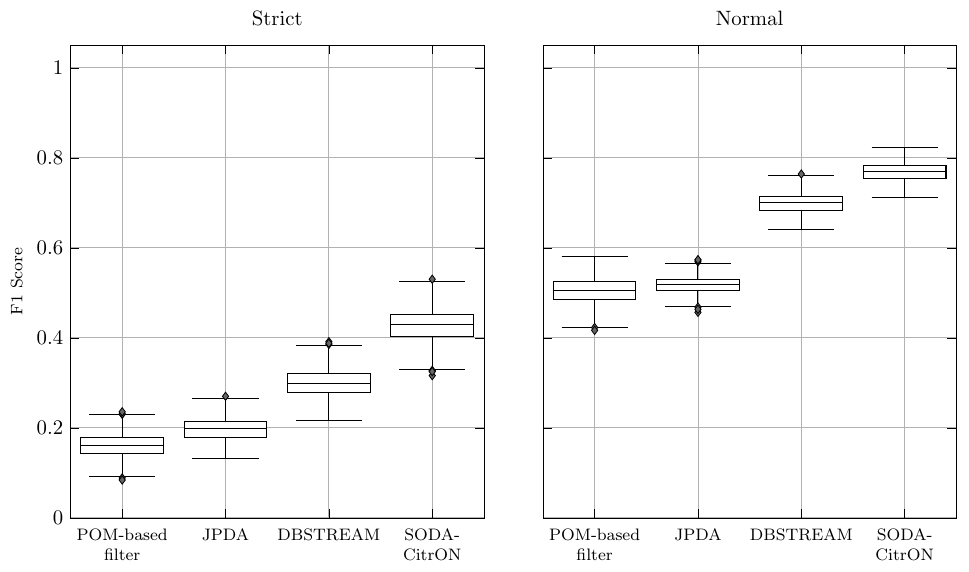}
         \caption{F1 score scenario B}
         \label{fig:F1 B}
     \end{subfigure}
     \hfill
     \begin{subfigure}[b]{0.48\textwidth}
         \centering
         \includegraphics[width=\linewidth, height=4cm]{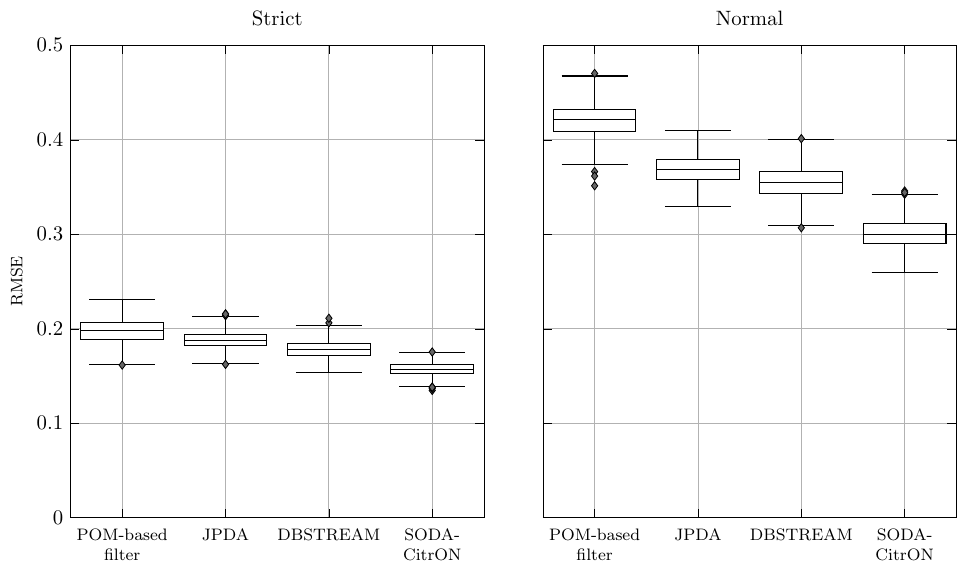}
         \caption{RMSE scenario B}
         \label{fig:RMSE B}
     \end{subfigure}
     
     \caption{Comparison of the key metrics for the different methods in both scenarios.}
     \label{fig:metric boxplots}
\end{figure*}
Fig. \ref{fig:metric boxplots} shows box plots for the F1 Score and the RMSE, where SODA-Citron consistently outperforms the other algorithms. To verify this, a Wilcoxon signed-rank test \cite{wilcoxon} was performed to compare SODA-Citron with all other methods, and for each test a very high significance level of $p<10^{-6}$ was found.
\begin{figure*}
    \centering
    \begin{subfigure}[b]{0.3\textwidth}
         \centering
         \includegraphics[width=\linewidth, height=3.5cm]{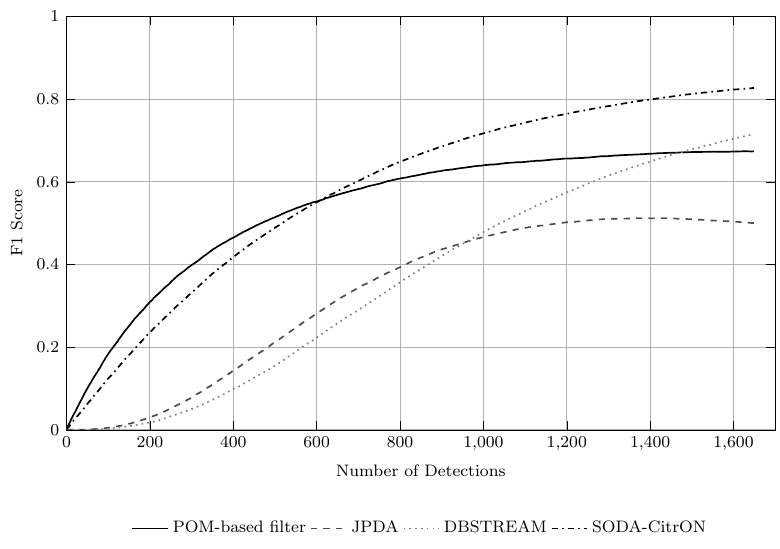}
         \caption{F1 Score over the number of detections}
         \label{fig:Online F1}
    \end{subfigure}
    \begin{subfigure}[b]{0.3\textwidth}
         \centering
         \includegraphics[width=\linewidth, height=3.5cm]{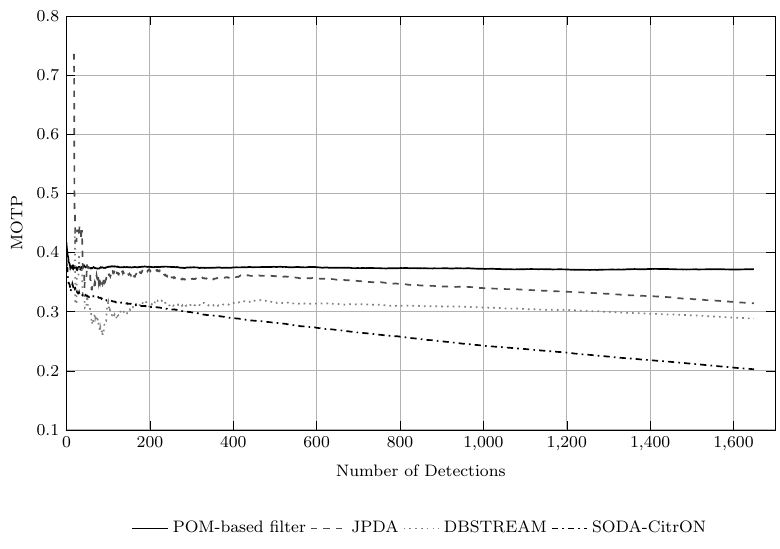}
         \caption{MOTP over the number of detections}
         \label{fig:Online MOTP}
    \end{subfigure}
    \begin{subfigure}[b]{0.3\textwidth}
         \centering
         \includegraphics[width=\linewidth, height=3.5cm]{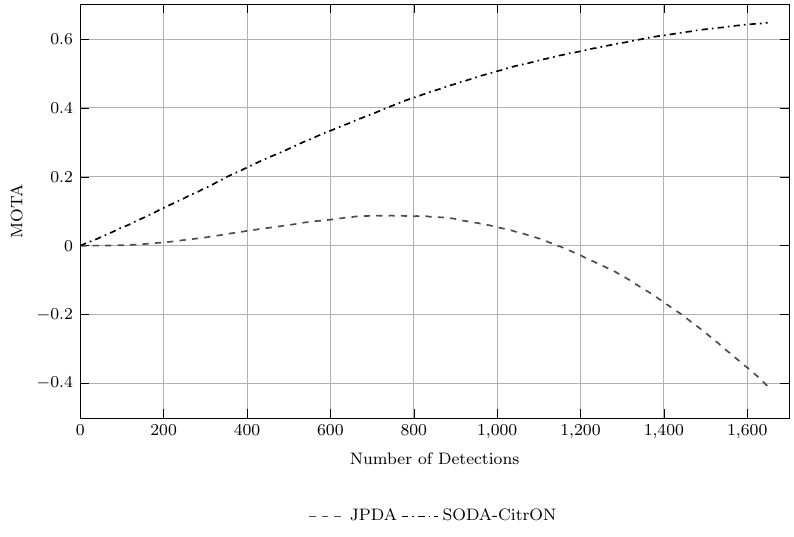}
         \caption{MOTA over the number of detections}
         \label{fig:Online RMSE}
    \end{subfigure}
    
     \caption{Comparison of the key online metrics for the different methods in scenario A.}
     \label{fig:online Metrics}
\end{figure*}
Fig. \ref{fig:online Metrics} shows the evolution of the F1 score over the number of detections, as well as the MOTP and MOTA tracking metrics based on the normal detection radius. The F1 score steadily increases with the number of detections for DBSTREAM and SODA-CitrON, while it plateaus or even declines for the POM-based filter and JPDA respectively. For MOTP, SODA-CitrON and JPDA are the algorithms that keep improving as more detections are added. Finally, MOTA keeps increasing for SODA-CitrON, suggesting a non-significant amount of object ID switches, while it even decreases for JPDA as a result of the higher number of false positives. The results for Scenario B showed a very similar algorithmic behavior, and therefore, the plots were not included.

\subsection{Discussion}
SODA-CitrON performed best on all analyzed metrics in this somewhat curated example, demonstrating its validity for online fusion and tracking for static objects in cluttered multi-modal environments. The use of detection confidence and sensor positioning uncertainty within SODA-CitrON gives it a clear advantage over existing methods. The confidence based weighting of SODA-CitrON helps in assigning targets faster when the confidence is high, while also reliably filtering clutter detections. In our experiment, DBSTREAM required three detections in a cluster to yield an output object (based on the parameterization), while SODA-CitrON produced a detection for a single observation, provided the confidence is high enough. Conversely, DBSTREAM will always assign an object, provided there are enough detections, while SODA-CitrON can still refuse assigning low confidence detections as a cluster. The parameter $w_{min}$ and the choice of $f(x)$ provide more control to end users in balancing precision and recall. Compared to JPDA, it does not suffer from accumulation of noisy estimations from clutter, as the confidence requirements remove noisy detections. This works only under the assumption that the clutter is spare or has low confidence. However, no method is expected to ever fully be capable of dealing with false sensor information. Using the uncertainty of the sensor position estimate with the information filter helps SODA-CitrON achieve a continual improvement of the position estimate as more detections are included. Compared to DBSTREAM, which uses the mean of the cluster, SODA-CitrON gives more weight to more precise detections, thus improving positional accuracy. 

Finally, a significant performance gain was shown, as SODA-CitrON proved to be more than 5 times faster than the second fastest method, DBSTREAM. Note that all implementations were done in Python, and for DBSTREAM and JPDA, third party implementations have been used that might not have been optimized for minimum runtime. However, a processing capability of $\sim250$ detections per second as observed on a single core makes SODA-CitrON viable for demanding real-time multi-sensor systems. 

\section{Conclusion}\label{sec:conclusion}

This paper addressed the problem of online fusion and data association of static objects from heterogeneous and temporally uncorrelated sensor detections. Although classical methods, such as JPDA, are effective for dynamic targets, their performance degrades in static object mapping scenarios with intermittent observations and heterogeneous uncertainties. To overcome these limitations, we proposed SODA-CitrON, a novel fully online, unsupervised approach based on clustering multi-modal sensor detections while simultaneously estimating positions and maintaining persistent tracks for an unknown number of objects.

The proposed method operates with worst-case loglinear complexity in the number of detections, making it suitable for real-time and large-scale deployments. Through extensive Monte Carlo simulations, we demonstrated that SODA-CitrON consistently outperforms state-of-the-art methods, including POM-based filtering, DBSTREAM clustering and JPDA, particularly in terms of positioning accuracy, F1-score, MOTP, and MOTA for various static object data association scenarios. These results highlight the robustness of the method in challenging multi-sensor scenarios with heterogeneous detection confidence and position uncertainty, as well as frequent clutter.

\subsection{Future Work}

Future studies should focus on testing SODA-CitrON in real-world scenarios with datasets obtained from field trials. In security related fields such as search \& rescue, humanitarian de-mining or CBRNE threat mapping, the proposed methodology could improve system capabilities \& performance. Furthermore, the influence of the method parameters on the data association performance must be systematically studied. In this context, it should be explored whether parameter optimization is viable for improving performance. Additionally, it should be investigated how the method can be generalized to work simultaneously for static and moving objects. Finally, more thorough experiments and data studies could be employed to formulate sensor requirements for successful static object data association.


\section*{Acknowledgment}
The authors used GPT-5 (OpenAI) for grammar and language editing assistance.


\bibliographystyle{IEEEtran}
\bibliography{references}

@book{bar1990tracking,
  title={Tracking and data association},
  author={Bar-Shalom, Yaakov and Fortmann, Thomas E and Cable, Peter G},
  year={1988},
  publisher={Academic Press, Inc.}
}

@INPROCEEDINGS{8999712,
  author={Xu, Shuoyuan and Shin, Hyo-Sang and Tsourdos, Antonios},
  booktitle={2019 Workshop on Research, Education and Development of Unmanned Aerial Systems (RED UAS)}, 
  title={Distributed Multi-Target Tracking with D-DBSCAN Clustering}, 
  year={2019},
  volume={},
  number={},
  pages={148-155},
  doi={10.1109/REDUAS47371.2019.8999712}}

@article{schraml2022real,
  title={Real-time gamma radioactive source localization by data fusion of 3d-lidar terrain scan and radiation data from semi-autonomous uav flights},
  author={Schraml, Stephan and Hubner, Michael and Taupe, Philip and Hofst{\"a}tter, Michael and Amon, Philipp and Rothbacher, Dieter},
  journal={Sensors},
  volume={22},
  number={23},
  pages={9198},
  year={2022},
  publisher={MDPI}
}

@Article{s21175715,
AUTHOR = {Kim, Jongwon and Cho, Jeongho},
TITLE = {DBSCAN-Based Tracklet Association Annealer for Advanced Multi-Object Tracking},
JOURNAL = {Sensors},
VOLUME = {21},
YEAR = {2021},
NUMBER = {17},
ARTICLE-NUMBER = {5715},
PubMedID = {34502605},
ISSN = {1424-8220},
DOI = {10.3390/s21175715}
}

@article{nurfalah_effective_2024,
	title = {Effective \& near real-time track-to-track association for large sensor data in {Maritime} {Tactical} {Data} {System}},
	volume = {10},
	issn = {2405-9595},
	doi = {10.1016/j.icte.2023.07.010},
	number = {2},
	urldate = {2026-02-02},
	journal = {ICT Express},
	author = {Nurfalah, Adiyasa and Supangkat, Suhono Harso and Mulyana, Eueung},
	month = apr,
	year = {2024},
	pages = {312--319},
}

@inproceedings{hubner2024bayesian,
  title={A Bayesian Approach-Data fusion for robust detection of vandalism and trespassing related events in the context of railway security},
  author={Hubner, Michael and Wohlleben, Kilian and Litzenberger, Martin and Veigl, Stephan and Opitz, Andreas and Grebien, Stefan and Dvorak, Maria-Theresia},
  booktitle={2024 27th International Conference on Information Fusion (FUSION)},
  pages={1--7},
  year={2024},
  organization={IEEE}
}

@article{hroob_adaptive_2024,
	title = {Adaptive robot localization in dynamic environments through self-learnt long-term {3D} stable points segmentation},
	volume = {181},
	issn = {0921-8890},
	doi = {10.1016/j.robot.2024.104786},
	urldate = {2026-01-29},
	journal = {Robotics and Autonomous Systems},
	author = {Hroob, Ibrahim and Molina, Sergi and Polvara, Riccardo and Cielniak, Grzegorz and Hanheide, Marc},
	month = nov,
	year = {2024},
	pages = {104786},
}

@article{shi_radar_2026,
	title = {Radar and {Camera} {Fusion} for {Object} {Detection} and {Tracking}: {A} {Comprehensive} {Survey}},
	volume = {28},
	issn = {1553-877X},
	shorttitle = {Radar and {Camera} {Fusion} for {Object} {Detection} and {Tracking}},
	doi = {10.1109/COMST.2025.3599596},
	urldate = {2026-01-29},
	journal = {IEEE Communications Surveys \& Tutorials},
	author = {Shi, Kun and He, Shibo and Shi, Zhenyu and Chen, Anjun and Xiong, Zehui and Chen, Jiming and Luo, Jun},
	year = {2026},
	pages = {3478--3520},
}

@article{samadzadegan_critical_2025,
	title = {A critical review on multi-sensor and multi-platform remote sensing data fusion approaches: current status and prospects},
	volume = {46},
	issn = {0143-1161},
	shorttitle = {A critical review on multi-sensor and multi-platform remote sensing data fusion approaches},
	doi = {10.1080/01431161.2024.2429784},
	number = {3},
	urldate = {2026-01-29},
	journal = {International Journal of Remote Sensing},
	publisher = {Taylor \& Francis},
	author = {Samadzadegan, Farhad and Toosi, Ahmad and Dadrass Javan, Farzaneh},
	month = feb,
	year = {2025},
	pages = {1327--1402},
}

@article{lekhak_viability_2024,
	title = {Viability of {Substituting} {Handheld} {Metal} {Detectors} with an {Airborne} {Metal} {Detection} {System} for {Landmine} and {Unexploded} {Ordnance} {Detection}},
	volume = {16},
	copyright = {http://creativecommons.org/licenses/by/3.0/},
	issn = {2072-4292},
	doi = {10.3390/rs16244732},
	language = {en},
	number = {24},
	urldate = {2026-01-29},
	journal = {Remote Sensing},
	publisher = {Multidisciplinary Digital Publishing Institute},
	author = {Lekhak, Sagar and Ientilucci, Emmett J. and Brinkley, Anthony Wayne},
	month = jan,
	year = {2024},
	pages = {4732},
}

@article{hasselmann_multi-robot_2024,
  title={A multi-robot system for the detection of explosive devices},
  author={Hasselmann, Ken and Malizia, Mario and Caballero, Rafael and Polisano, Fabio and Govindaraj, Shashank and Stigler, Jakob and Ilchenko, Oleksii and Bajic, Milan and De Cubber, Geert},
  journal={arXiv preprint arXiv:2404.14167},
  year={2024}
}

@article{bernardin2008evaluating,
  title={Evaluating multiple object tracking performance: the clear mot metrics},
  author={Bernardin, Keni and Stiefelhagen, Rainer},
  journal={EURASIP Journal on Image and Video Processing},
  volume={2008},
  number={1},
  pages={246309},
  year={2008},
  publisher={Springer}
}

@article{hahsler_clustering_2016,
	title = {Clustering {Data} {Streams} {Based} on {Shared} {Density} between {Micro}-{Clusters}},
	volume = {28},
	issn = {1558-2191},
	doi = {10.1109/TKDE.2016.2522412},
	number = {6},
	urldate = {2026-01-29},
	journal = {IEEE Transactions on Knowledge and Data Engineering},
	author = {Hahsler, Michael and Bolaños, Matthew},
	month = jun,
	year = {2016},
	pages = {1449--1461},
}

@inproceedings{cao_density-based_2006,
	title = {Density-{Based} {Clustering} over an {Evolving} {Data} {Stream} with {Noise}},
	volume = {2006},
	doi = {10.1137/1.9781611972764.29},
	author = {Cao, Feng and Ester, Martin and Qian, Weining and Zhou, Aoying},
	month = apr,
	year = {2006},
}

@inproceedings{ester_density-based_1996,
	title = {A density-based algorithm for discovering clusters in large spatial databases with noise},
	urldate = {2026-01-29},
	booktitle = {Proceedings of the {Second} {International} {Conference} on {Knowledge} {Discovery} and {Data} {Mining}},
	publisher = {AAAI Press},
	author = {Ester, Martin and Kriegel, Hans-Peter and Sander, Jörg and Xu, Xiaowei},
	month = aug,
	year = {1996},
	pages = {226--231},
}

@article{guttman_r-trees_1984,
	title = {R-trees: a dynamic index structure for spatial searching},
	volume = {14},
	issn = {0163-5808},
	shorttitle = {R-trees},
	doi = {10.1145/971697.602266},
	number = {2},
	urldate = {2026-01-29},
	journal = {SIGMOD Rec.},
	author = {Guttman, Antonin},
	month = jun,
	year = {1984},
	pages = {47--57},
}

@inproceedings{wohlleben_bayesian_2025,
	title = {Bayesian {Optimization} for {Parameter} {Selection} in {Fusion} {Systems}},
	doi = {10.23919/FUSION65864.2025.11124011},
	urldate = {2026-01-29},
	booktitle = {2025 28th {International} {Conference} on {Information} {Fusion} ({FUSION})},
	author = {Wohlleben, Kilian and Siems, Finn and Nausner, Jan and Hubner, Michael},
	month = jul,
	year = {2025},
	pages = {1--7},
}

@article{barker_static_1998,
	title = {Static data association with a terrain-based prior density},
	volume = {28},
	issn = {1558-2442},
	doi = {10.1109/5326.661097},
	number = {1},
	urldate = {2025-10-15},
	journal = {IEEE Transactions on Systems, Man, and Cybernetics, Part C (Applications and Reviews)},
	author = {Barker, A.L. and Brown, D.E. and Martin, W.N.},
	month = feb,
	year = {1998},
	pages = {151--157},
}

@inproceedings{guler_stationary_2007,
	title = {Stationary objects in multiple object tracking},
	doi = {10.1109/AVSS.2007.4425318},
	urldate = {2025-10-15},
	booktitle = {2007 {IEEE} {Conference} on {Advanced} {Video} and {Signal} {Based} {Surveillance}},
	author = {Guler, Sadiye and Silverstein, Jason A. and Pushee, Ian H.},
	month = sep,
	year = {2007},
	pages = {248--253},
}

@inproceedings{schueler_360_2012,
	title = {360 {Degree} multi sensor fusion for static and dynamic obstacles},
	doi = {10.1109/IVS.2012.6232253},
	urldate = {2025-10-15},
	booktitle = {2012 {IEEE} {Intelligent} {Vehicles} {Symposium}},
	author = {Schueler, Kai and Weiherer, Tobias and Bouzouraa, Essayed and Hofmann, Ulrich},
	month = jun,
	year = {2012},
	pages = {692--697},
}

@inproceedings{bowman_probabilistic_2017,
	title = {Probabilistic data association for semantic {SLAM}},
	doi = {10.1109/ICRA.2017.7989203},
	urldate = {2025-10-16},
	booktitle = {2017 {IEEE} {International} {Conference} on {Robotics} and {Automation} ({ICRA})},
	author = {Bowman, Sean L. and Atanasov, Nikolay and Daniilidis, Kostas and Pappas, George J.},
	month = may,
	year = {2017},
	pages = {1722--1729},
}

@book{bar2001estimation,
  title={Estimation with applications to tracking and navigation: theory algorithms and software},
  author={Bar-Shalom, Yaakov and Li, X Rong and Kirubarajan, Thiagalingam},
  year={2001},
  publisher={John Wiley \& Sons}
}

@INPROCEEDINGS{10224185,
  author={Hiscocks, Steven and Barr, Jordi and Perree, Nicola and Wright, James and Pritchett, Henry and Rosoman, Oliver and Harris, Michael and Gorman, Roisín and Pike, Sam and Carniglia, Peter and Vladimirov, Lyudmil and Oakes, Benedict},
  booktitle={2023 26th International Conference on Information Fusion (FUSION)}, 
  title={Stone Soup: No Longer Just an Appetiser}, 
  year={2023},
  volume={},
  number={},
  pages={1-8},
  doi={10.23919/FUSION52260.2023.10224185}
}

@misc{montiel_river_2020,
	title = {River: machine learning for streaming data in {Python}},
	shorttitle = {River},
	doi = {10.48550/arXiv.2012.04740},
	urldate = {2026-01-29},
	publisher = {arXiv},
	author = {Montiel, Jacob and Halford, Max and Mastelini, Saulo Martiello and Bolmier, Geoffrey and Sourty, Raphael and Vaysse, Robin and Zouitine, Adil and Gomes, Heitor Murilo and Read, Jesse and Abdessalem, Talel and Bifet, Albert},
	month = dec,
	year = {2020},
}

@book{blackman1999design,
  title={Design and analysis of modern tracking systems},
  language = {en},
  publisher = {Artech House},
  author={Blackman, Samuel S and Popoli, Robert},
  year={1999},
}

@article{wilcoxon,
 ISSN = {00994987},
 author = {Frank Wilcoxon},
 journal = {Biometrics Bulletin},
 number = {6},
 pages = {80--83},
 publisher = {[International Biometric Society, Wiley]},
 title = {Individual Comparisons by Ranking Methods},
 urldate = {2026-02-09},
 volume = {1},
 year = {1945}
}

\end{document}